\documentclass[twoside,11pt]{article}
\pdfoutput=1

\usepackage{geometry}
\usepackage{natbib}
\usepackage{graphicx}
\usepackage{rotating}
\usepackage{amsmath}
\usepackage{amssymb}

\usepackage{theorem}
\newcommand{\BlackBox}{\rule{1.5ex}{1.5ex}}  

\newtheorem{theorem}{Theorem}

\newcommand{\cbr}[1]{\left\{#1\right\}}

\DeclareMathOperator*{\argmin}{\mathrm{argmin}}

\usepackage{multicol}
\usepackage{enumerate,mdwlist}

\newcommand{\intset}[1]{\cbr{1..n}}

\usepackage[colorlinks,pdfpagelabels,plainpages=false]{hyperref}
\usepackage{algorithm}
\usepackage{rotating}
\usepackage{algpseudocode}

\usepackage{xcolor}
\definecolor{dark-red}{rgb}{0.4,0.15,0.15}
\definecolor{dark-blue}{rgb}{0.15,0.15,0.4}
\definecolor{medium-blue}{rgb}{0,0,0.5}
\hypersetup{
   colorlinks, linkcolor={dark-blue},
   citecolor={dark-blue}, urlcolor={medium-blue}
}

\usepackage{booktabs}
\usepackage{bm}
\usepackage{amsmath}
\usepackage{amssymb}
\usepackage{amsfonts}
\usepackage{mathtools}
\usepackage{comment}
\usepackage{subfigure}

\usepackage{color}

\usepackage{array}
\newcolumntype{L}[1]{>{\raggedright\let\newline\\\arraybackslash\hspace{0pt}}m{#1}}
\newcolumntype{C}[1]{>{\centering\let\newline\\\arraybackslash\hspace{0pt}}m{#1}}
\newcolumntype{R}[1]{>{\raggedleft\let\newline\\\arraybackslash\hspace{0pt}}m{#1}}

\usepackage{parskip}

\title{Lass\texttt{0}: sparse non-convex regression by local search}

\author{
  William Herlands \\
  Carnegie Mellon University \\
  herlands@cmu.edu
  \and
  Maria De-Arteaga \\
  Carnegie Mellon University \\
  mdeartea@andrew.cmu.edu
  \and
  Daniel Neill \\
  Carnegie Mellon University \\
  neill@cs.cmu.edu
  \and
  Artur Dubrawski \\
  Carnegie Mellon University \\
  awd@cs.cmu.edu
}

\begin{document}
\date{}  

\maketitle

\begin{abstract} 

\begin{sloppypar}
We compute approximate solutions to $L_0$ regularized linear regression using $L_1$ regularization, also known as the Lasso, as an initialization step. Our algorithm, the Lass\texttt{0} (``Lass-zero''), uses a computationally efficient stepwise search to determine a locally optimal $L_0$ solution given any $L_1$ regularization solution. We present theoretical results of consistency under orthogonality and appropriate handling of redundant features. Empirically, we use synthetic data to demonstrate that Lass\texttt{0} solutions are closer to the true sparse support than $L_1$ regularization models. Additionally, in real-world data Lass\texttt{0} finds more parsimonious solutions than $L_1$ regularization while maintaining similar predictive accuracy.
\end{sloppypar}

\end{abstract} 

\section{Introduction}
\label{sec:intro}
Sparse approximate solutions to linear systems are desirable for providing interpretable results that succinctly identify important features. For  $X \in \mathbb{R}^{n \times p}$ and $y \in \mathbb{R}^{n}$, $L_0$ regularization (Eq. \ref{eq:L0_reg}\footnote{We consider the Lagrange form of subset selection. Since the problem is nonconvex this is weaker than the constrained form, meaning that all solutions of the Lagrange problem are solutions to a constrained problem.}), called ``best subset selection,'' is a natural way to achieve sparsity by directly penalizing non-zero elements of $\beta$.
This intuition is fortified by theoretical justification. \cite{FG1994} demonstrate that for general predictor matrix $X$, $L_0$ regularization achieves the asymptotic minimax rate of risk inflation. Unfortunately, it is well known that $L_0$ regularization is non-convex and NP hard \citep{Balas1995}.
\vspace{-3em}
\begin{multicols}{2}
\begin{eqnarray}
\label{eq:L0_reg}
\min_{\beta \in R^p} \frac{1}{2}\| y - X\beta \|_2^2 + \lambda\|\beta\|_0
\end{eqnarray}
\break
\begin{eqnarray}
\label{eq:L1_reg}
\min_{\beta \in R^p} \frac{1}{2}\| y - X\beta \|_2^2 + \lambda\|\beta\|_1
\end{eqnarray}
\end{multicols}
Despite the computational difficulty, the optimality of $L_0$ regularization has motivated approximation methods such as \cite{zhang2009}, who provide a Forward-Backward greedy algorithm with asymptotic guarantees. Additionally, integer programming has been used to find solutions for problems of bounded size \citep{konno2009choosing,miyashiro2015subset,gatu2012branch}.

Instead of $L_0$ regularization, it is common to use $L_1$ regularization (Eq. \ref{eq:L1_reg}), known as the Lasso \citep{Tib1996}. This convex relaxation of $L_0$ regularization achieves sparse solutions which are model consistent and unique under regularity conditions, which, among other things, limit the correlations between columns of $X$ \citep{zhao2006model, tibshirani2013lasso}. Additionally, $L_1$ is a reasonable substitute for $L_0$ regularization because the $L_1$ norm is the best convex approximation to the $L_0$ norm \citep{RKA2012}. However, on real-world data sets, $L_1$ regularization tends to select incorrect models since the $L_1$ norm shrinks all coefficients including those which are in the active set \citep{friedman2012fast, mazumder2011sparsenet}. This bias can be particularly troublesome in very sparse settings, where the predictive risk of $L_1$ can be arbitrarily worse than that of $L_0$ \citep{LPFU2008}. 

In order to take advantage of the computational tractability of $L_1$ regularization, and the optimality of $L_0$, we develop the Lass\texttt{0} (``Lass-zero''), a method which uses an $L_1$ solution as an initialization step to find a locally optimal $L_0$ solution. At each computationally efficient step, the Lass\texttt{0} improves upon the $L_0$ objective, often finding sparser solutions without sacrificing prediction accuracy. 

Previous literature, such as SparseNet \citep{mazumder2011sparsenet}, also explored the relationship between $L_1$ and $L_0$ solutions. Yet unlike our approach, SparseNet reparameterizes the problem with MC+ loss and solves a generalized soft thresholding problem at each iteration requiring a large number of problems to solve to reach $L_0$. Alternatively, \cite{johnson2015risk} use the $L_0$ objective as a criterion to select among different $L_1$ models from the LARS \citep{EHJT2004} solution set. Additionally, they compare forward stepwise $L_0$ regression to $L_1$ regression. However, in neither case do they improve upon the $L_1$ results by optimizing $L_0$ directly, as in our work.

In the remainder of this paper, Section \ref{sec:lass0} details the  Lass\texttt{0} algorithm. Section \ref{sec:theoretical} provides theoretical guarantees for convergence in the orthogonal case and elimination of redundant variables in the general case. Section \ref{sec:experimental} presents empirical results on synthetic and real world data. Finally, we conclude in Section \ref{sec:conclusions} with directions for future work in the general context of non-convex optimization.

\section{Lass\texttt{0}}
\label{sec:lass0}

We propose a new method for finding sparse approximate solutions to linear problems, which we call the Lass\texttt{0}. The full pseudocode of the Lass\texttt{0} algorithm is presented in Algorithm \ref{CHalgorithm}, and we refer to the lines through this section. The method is initialized by a solution to L$_1$ regularization, $\beta^{L_1}$, given a particular $\lambda$. The Lass\texttt{0} then uses an iterative algorithm to find a locally optimal solution that minimizes the objective function of the L$_0$ regularization (Eq. \ref{eq:L0_objective}). 
\begin{eqnarray}
\label{eq:L0_objective}
L_0(\beta, y, X, \lambda) = \| y - X\beta \|^2_2 + \lambda\|\beta\|_0
\end{eqnarray}
If $supp()$ indicates the support, the first step in the Lass\texttt{0} is to compute $\beta=\hat{OLS}(supp(\beta^{L_1}),y,X)$, the ordinary least squares solution constrained such that every zero entry of $\beta^{L_1}$ must remain zero. $\hat{OLS}()$ is formally defined as,
\begin{eqnarray}
\hat{OLS}(F,y,X) = \min_{\beta} \| y - X\beta \|^2_2 \mbox { s.t. } \beta_i= 0 \mbox{ } \forall i \notin F
\end{eqnarray}
For each entry, $\beta_i$, of the resulting vector, we compute the effect of individually adding or removing it from $supp(\beta)$ in Lines 6 and 7. Note that by adding an entry to the support, we increase the penalty by $\lambda$, but potentially create a better estimate for $y$, resulting in a lower $\| y - X\beta \|_2^2$ loss term. Similarly, the opposite may be true when removing an entry from the support set.

This procedure yields a new solution vector $\beta^{(i)}$ for each $i$. The $\beta^{(i)}$ which minimizes the $L_0$ objective function is selected as $\beta'$ in Line 8. Then, in Line 9, we accept $\beta'$ only if it is strictly better than the solution we began with, $\beta$. The iterative algorithm terminates whenever there is no improvement.
\begin{algorithm}
\caption{Lass\texttt{0}}
\label{CHalgorithm}
\begin{algorithmic}[1]
\State Input: $L_1$ solution, $\beta^{L_1}$
\State $F =supp(\beta^{L_1})$
\State $\beta = \hat{OLS}(F)$
\While{True} 
 \State $F =supp(\beta_i)$ 
 \State \textbf{For all} $i \in F$ \textbf{do:} $\beta^{(i)} = \hat{OLS}(F \setminus \{i\}, y, X)$
 \State \textbf{For all} $i \notin F$ \textbf{do:} $\beta^{(i)} = \hat{OLS}(F \cup  \{i\}, y, X)$
 \State $\beta' = \argmin_{i}\limits L_0(\beta^{(i)}, y, X, \lambda)$
 \If{$L_0(\beta', y, X, \lambda) < L_0(\beta, y, X, \lambda)$}
 	\State $\beta = \beta'$
 \Else
 	Break
 \EndIf
\EndWhile
\end{algorithmic}
\end{algorithm}

This procedure is equivalent to greedy coordinate minimization where we warm-start the optimization procedure with the $L_1$ regularization solution. Additionally, we note that any $L_p$ regularization with norm $p<1$ is non-convex. While the present work focuses on $L_0$ regularization, the Lass\texttt{0} can be applied to approximate solutions to any other non-convex $L_p$ regularization with minimal changes.

\section{Theoretical properties}
\label{sec:theoretical}

\begin{theorem}
Assuming that $X$ is orthogonal, the Lass\texttt{0} solution is the $L_0$ regularization solution.
\end{theorem}

\textit{Proof.} Recall that Lass\texttt{0} is initialized with the $L_1$ regularization solution. With an orthogonal set of covariates, it is well known that the solution to $L_1$ regularization, $\beta^{L_1}$, is soft-thresholding of the components of $X^Ty$ at level $\lambda$ (Eq. \ref{eq:soft_th}). Additionally, it is well known that in this case the solution to L$_0$ regularization, $\beta^{L_0}$, is hard-thresholding of the components of $X^ty$ at level $\sqrt{2\lambda}$ (Eq. \ref{eq:hard_th}).
\vspace{-2em}
\begin{multicols}{2}
\begin{eqnarray}
\beta^{L_1}_j=\left\{ \begin{array}{rcl}
X_j^Ty - \lambda & \mbox{if} & X_j^Ty > \lambda \\
0 & \mbox{if} & |X_j^Ty|< \lambda \\
X_j^Ty + \lambda & \mbox{if} &  X_j^Ty < -\lambda
\end{array}\right.
\label{eq:soft_th}
\end{eqnarray}
\break
\vspace{-1.5em}
\begin{eqnarray}
\beta^{L_0}_j=\left\{ \begin{array}{rcl}
X_j^Ty & \mbox{if} & X_j^Ty>\sqrt{2\lambda} \\
0 & \mbox{if} & |X_j^Ty|< \sqrt{2\lambda} \\
X_j^Ty & \mbox{if} &  X_j^Ty < -\sqrt{2\lambda}
\end{array}\right.
\label{eq:hard_th}
\end{eqnarray}
\end{multicols}
We will prove that the Lass\texttt{0} solution, $\beta^{Lass\texttt{0}} = \beta^{L_0}$. Since the solutions to $L_1$ and $L_0$ regularization depend on $\lambda$, the proof is divided in three cases to cover all possible values of $\lambda$, and we use the same regularization parameter $\lambda$ for both algorithms.

\begin{enumerate}[i.]

\item \textbf{Case }$\pmb{\lambda=2}$\textbf{:} Since $\sqrt{2\lambda}=\lambda$, therefore $supp(\beta^{L_0}) = supp(\beta^{L_1})$. Note that in the orthogonal case, the least squares solution is $(X^TX)^{-1}X^ty=X^ty$. In the first step of the Lass\texttt{0} algorithm we find $\hat{OLS}(supp(\beta^{L_1}),y,X)$ which corresponds to setting $\beta_k = (X^Ty)_k \mbox{ } y \mbox{  } \forall k \in supp(\beta^{L_1})$, and $\beta_k=0$ otherwise. Therefore, in the first step the algorithm will reach the hard-thresholding at level $\sqrt{2\lambda}$ and terminate.

\item \textbf{Case }$\pmb{\lambda>2}$\textbf{:} Since $\sqrt{2\lambda} < \lambda$, then $supp(\beta^{L_0}) \supseteq supp(\beta^{L_1})$. Let $\beta=\hat{OLS}(supp(\beta^{L_1}), y, X)$ and let $\beta^{new} = \hat{OLS}(supp(\beta) \setminus \{k\}, y, X)$. The Lass\texttt{0} will only choose $\beta^{new}$ and remove element $k$ from $supp(\beta^{L_1})$ if,
\begin{eqnarray}
\frac{1}{2}\|y-X\beta^{new}\|_2^2 - \frac{1}{2}\|y-X\beta\|_2^2 < \lambda
\label{orthog_ii}
\end{eqnarray}

Yet such inequality will never hold, since $\beta_k = (X^Ty)_k$ and it would imply 
\begin{eqnarray}
\beta_k(X^Ty)_k - \frac{1}{2} (\beta_k)^2 < \lambda \hspace{0.2in}
\Rightarrow \hspace{0.2in} \frac{1}{2} (X^Ty)^2_k < \lambda
 \hspace{0.2in}
\Rightarrow  \hspace{0.2in} \frac{1}{2} \lambda^2 < \lambda
\end{eqnarray}
Which contradicts $\lambda > 2$. Thus Lass\texttt{0} will never remove an element from $supp(\beta^{L_1})$.

Similarly, if we let $\beta^{new} = \hat{OLS}(supp(\beta) \cup \{k\}, y, X)$ Lass\texttt{0} will only choose $\beta^{new}$ and add element $k$ to $supp(\beta^{L_1})$ if,
\begin{eqnarray}
\label{eq:ortho_two_two}
\frac{1}{2}\|y-X\beta^{new}\|_2^2 - \frac{1}{2}\|y-X\beta\|_2^2 < -\lambda \hspace{0.2in}
\Rightarrow \hspace{0.2in} -\frac{1}{2} (X^Ty)^2_k < -\lambda
\label{orthog_ii_2}
\end{eqnarray}
Thus Lass\texttt{0} will add element $k$ to $supp(\beta^{L_1})$ if and only if $\sqrt{2\lambda} < (X^Ty)_k$. Therefore, $supp(\beta^{Lass\texttt{0}}) = supp(\beta^{L_0})$. Furthermore, since $\beta^{Lass\texttt{0}}$ is optimized by OLS, $\beta^{Lass\texttt{0}} = \beta^{L_0}$.

\item \textbf{Case }$\pmb{\lambda<2}$\textbf{:} Since $\sqrt{2\lambda} > \lambda$, then $supp(\beta^{L_0}) \subseteq supp(\beta^{L_1})$. The result that $\beta^{Lass\texttt{0}} = \beta^{L_0}$ follows from an analogous argument to the above, omitted for the sake of brevity. $\blacksquare$

\end{enumerate}

For sparse solutions, it is important to know how a given algorithm will behave when faced with strongly correlated features. For example, the elastic net \citep{zou2005regularization} assigns identical coefficients to identical variables. In contrast, $L_1$ regularization picks one of the strongly correlated features. The latter behavior is desirable in situations where including both variables in the support would be considered redundant. We now prove that when two variables are strongly correlated, Lass\texttt{0} behaves similarly to $L_1$ regularization: it only selects one among a group of strongly correlated features. 

\begin{theorem}
Assume that $\mathbf{x_i}=k\mathbf{  x_j}$, then either $\beta_i^{Lass\texttt{0}}=0$ or $\beta_j^{Lass\texttt{0}}=0$ (or both).
\end{theorem}

\textit{Proof.} Let $\beta$ be the solution at any step of Lass\texttt{0}. We will prove that if both indices $\{i,j\}\in supp(\beta)$, meaning $\beta_i \neq 0$ and $\beta_j \neq 0$, at least one of them will become zero in the solution.

Without loss of generality, let $\beta^{new}$ be the least squares solution that preserves all the constraints of $\beta$ and also enforces $\beta^{new}_i=0$. Let $\beta^{new}_j = k\beta_i + \beta_j$, then $||y-X\beta^{new}||_2^2=||y-X\beta||_2^2$, implying $L_0(\beta_{new},y,X,\lambda)<L_0(\beta,y,X,\lambda)$. $\blacksquare$

\section{Experimental results}
\label{sec:experimental}

\begin{figure}
\begin{center}
\includegraphics[width=1.0\linewidth]{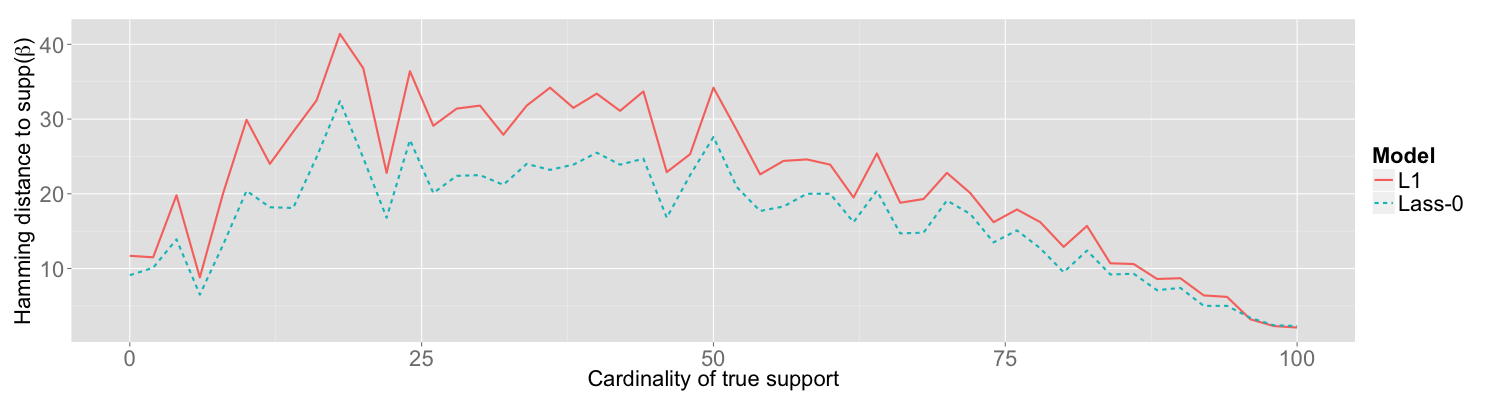}
\caption{Average Hamming distance between $supp(\beta^{Lass\texttt{0}})$ or $supp(\beta^{L_1})$ and the true $supp(\beta)$ over 10 CV tests. The Lass\texttt{0} consistently chooses models closer to the true support.}
\label{fig:ErrorsSupport}
\end{center}
\end{figure}

We generate synthetic data from a linear model $y = X\beta + \epsilon$, where each sample is generated $X_j\sim N(\mu, \Sigma)$ using $\Sigma$ with high correlation. The coefficients are generated as $\beta \sim \text{Uniform}(-1,1)$, with sparsity enforced by setting some $\beta_i$ to zero. We compare $supp(\beta^{Lass\texttt{0}})$ and $supp(\beta^{L_1})$ against the true underlying support, $supp(\beta)$. We use 10-fold cross validation (CV) testing and report the average Hamming distance between the estimated and true supports. Figure \ref{fig:ErrorsSupport} shows Hamming distances over different levels of sparsity in the true support. The Lass\texttt{0} consistently yields models which are closer to the true support than the optimally chosen $L_1$ model.

We evaluate the Lass\texttt{0} on nine real-world data sets sourced from the publicly available repositories \citep{valdar2006genome,Lic2013}. Table \ref{tab:real_data_results} shows the mean and standard deviation for the normalized root mean squared error (NRMSE) and cardinality of the support for the estimated $\beta$. For all data sets, both regularization methods produce very similar NRMSE values. However, in most cases the Lass\texttt{0} reduced the size of the active set, often by $50\%$ or more. Combined with the above results showing that the Lass\texttt{0} yields models closer to the true sparse synthetic model, we see that the Lass\texttt{0} tends to produce sparser, more fidelitous models than $L_1$ regularization. 
\begin{table}[]
\centering
\begin{tabular}{l|l|l|l|l|l}
Data          & NRMSE $L_1$            & NRMSE Lass\texttt{0}       & $p$ & $|supp(\beta^{L_1})|$   & $|supp(\beta^{Lass\texttt{0}})|$ \\ \hline
Pyrimidines   & 101.4 $\pm$  47.5 & 103.1 $\pm$ 42.3 & 28          & 16.1 $\pm$ 5.5 & 7.6 $\pm$ 5.1    \\
Ailerons      & 42.2 $\pm$ 1.8    & 42.5 $\pm$ 1.9   & 41          & 24 $\pm$ 3.3   & 6.9 $\pm$ 1.1    \\
Triazines     & 98.9 $\pm$ 14.5   & 97.5 $\pm$ 19.7  & 61          & 17.9 $\pm$ 9.2 & 7.3 $\pm$ 6.6    \\
Airplane stocks   & 36.1 $\pm$ 5.2    & 36.4 $\pm$ 5.6   & 10          & 8.5 $\pm$ 0.5  & 7.8 $\pm$ 0.9    \\
Pole Telecomm & 73.3 $\pm$ 2.1    & 73.4 $\pm$ 2.1   & 49          & 22.7 $\pm$ 1.1 & 24.5 $\pm$ 0.9   \\
Bank domains       & 69.9 $\pm$ 3      & 70.7 $\pm$ 3.1   & 33          & 9.2 $\pm$ 2.7  & 5.2 $\pm$ 8.2    \\
Pumadyn domains       & 89 $\pm$ 2.2      & 88.9 $\pm$ 2.4   & 33          & 5.2 $\pm$ 8.2  & 1 $\pm$ 0        \\
Breast Cancer & 93.5 $\pm$ 15.3   & 96.7 $\pm$ 19.7  & 33          & 16 $\pm$ 8.7   & 18.8 $\pm$ 5.7   \\
Mice          & 103.5 $\pm$  5    & 105 $\pm$ 6.6    & 100         & 17 $\pm$ 6.6   & 6.3 $\pm$ 4.3   
\end{tabular}
\caption{Mean and standard deviation from  Lass\texttt{0} and $L_1$ regularization on real data for 10 CV runs}
\label{tab:real_data_results}
\end{table}

\section{Future work}
\label{sec:conclusions}
We intend to build upon Theorem 1 to support our empirical observations. Additionally, we expect that this paper's general approach can be applied to other non-convex optimization problems. While convex relaxations may yield interesting problems in their own right, they are often good approximations to non-convex solutions. Using convex results to initialize an efficient search for a locally optimal non-convex solution can combine the strengths of convex and non-convex formulations.

\subsubsection*{Acknowledgments}

We thank Ryan Tibshirani for his advice and fruitful suggestions. This material is based upon work supported by the National Science Foundation Graduate Research Fellowship under Grant No. DGE 1252522 and the National Science Foundation award No. IIS-0953330.
\clearpage

\bibliographystyle{apalike}
\bibliography{bib}

\end{document}